\documentclass{article}

 \usepackage{arxiv}

\usepackage[numbers, compress]{natbib}
\usepackage[utf8]{inputenc}
\usepackage[T1]{fontenc}
\usepackage{hyperref}
\usepackage{url}
\usepackage{booktabs}
\usepackage{amsfonts}
\usepackage{nicefrac}
\usepackage{microtype}
\usepackage{xcolor}
\usepackage{bm}
\usepackage{amsmath,amssymb,amsthm}
\usepackage{mathtools}
\usepackage{multirow}
\usepackage{algorithm}
\usepackage{algorithmic}
\usepackage{enumitem}
\usepackage{makecell}

\newcommand{\zy}[1]{#1}
\usepackage[most]{tcolorbox}
\usepackage{xcolor}
\usepackage{amsmath, amssymb}

\definecolor{tplhead}{HTML}{4D4D4D}
\definecolor{tplbody}{HTML}{F7F7F7}

\newtcolorbox{promptbox}{
    colback=gray!3,
    colframe=black!55,
    boxrule=0.5pt,
    arc=2pt,
    left=10pt, right=10pt, top=8pt, bottom=8pt,
    enhanced,
}

\title{Enhancing Decision-Making with Large Language Models through Multi-Agent Fictitious Play}

\author{%
  Leyang Shen \\
  National University of Singapore \\
  Singapore \\
  \texttt{lshen@u.nus.edu} \\
  \And
  Yang Zhang\thanks{Corresponding author} \\
  National University of Singapore \\
  Singapore \\
  \texttt{zhangy@nus.edu.sg}
  \And
  Xiaoyan Zhao \\
  National University of Singapore \\
  Singapore \\
  \texttt{xzhao@se.cuhk.edu.hk} \\
  \And
  Chun Kai Ling \\
  National University of Singapore \\
  Singapore \\
  \texttt{chunkail@nus.edu.sg} \\
  \And
  Tat-Seng Chua \\
  National University of Singapore \\
  Singapore \\
  \texttt{dcscts@nus.edu.sg} \\
}

\begin{document}

\maketitle

\begin{abstract}

Large language model (LLM)-based multi-agent systems (MAS) have demonstrated great potential in solving tasks with \emph{execution complexity}, by distributing subtasks across cooperative agents. However, this divide-and-conquer paradigm falls short on decision-making tasks that are also prevalent in the real world. These tasks require simultaneous reasoning from the stances of all involved stakeholders whose decisions are mutually dependent and thus cannot be solved in isolation. 
We characterize this challenge as \emph{stance entanglement}, a form of decision complexity distinct from execution complexity. To address it, we propose Multi-Agent Fictitious Play (MAFP), a novel MAS paradigm that represents stakeholder stances as agents and formulates decision-making as an equilibrium-seeking process. Built on the game-theoretic principle of fictitious play, MAFP iteratively updates each agent's decision by best responding to the empirical mixture of other agents' past decisions. This enables agents to expose and address one another's weaknesses, progressively improving decision quality and robustness. We evaluate MAFP on challenging decision-making tasks that test the capability of deciding strategies for competitive scenarios prior to acting. MAFP outperforms both single-round and multi-round baselines on two complementary metrics, \emph{tournament strength} and \emph{robustness}, demonstrating its effectiveness in addressing stance entanglement.

\end{abstract}

\section{Introduction}

\begin{figure}[t]
    \centering
    \includegraphics[width=1.0\linewidth]{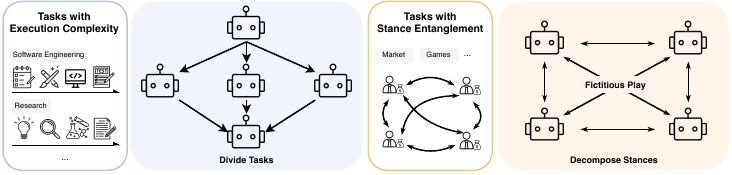}
    \caption{Existing MAS address execution complexity, as in software engineering or research (left), by dividing a task into subtasks across cooperative agents. In contrast, MAFP targets stance entanglement, as in competitive market or strategic games (right), where stakeholders' decisions are mutually dependent: it decomposes these entangled stances into agents and derives decisions through fictitious play.}
    \label{fig:intro}
\end{figure}

Large language model (LLM)-based multi-agent systems (MAS)~\cite{zhang2024chain, tran2025multi} have emerged as a powerful paradigm for solving complex tasks that exceed the capability of a single LLM call~\cite{wang2024agent-survey}, with applications in software engineering~\cite{deng2025swe-pro}, deep research~\cite{xu2025deepresearch}, and scientific discovery~\cite{zhao2026agentic}.
The core of these MAS is divide and conquer~\cite{gu2025agentgroupchat}, where each agent corresponds to a part of the task, accomplishing the task collaboratively.
By doing so, MAS reduces the difficulty faced by each individual agent and alleviates the context-window pressure that bottlenecks single-agent reasoning~\cite{team2026kimi}, thereby increasing the overall performance.

The complexity that existing MAS address is primarily \emph{execution complexity}: tasks are difficult because their execution requires long reasoning chains~\cite{lei2024macm}, broad information coverage~\cite{xu2025deepresearch, manus2025wideresearch}, or heterogeneous skills~\cite{zhao2026agentic}, thus these burdens can be distributed across cooperative agents, as demonstrated in Fig.~\ref{fig:intro} (Left).
However, many real-world decision-making tasks, like negotiation~\cite{bianchi2024negoarena}, game~\cite{duan2024gtbench}, and competitive market~\cite{backlund2025vending, zhang2026retailbench}, introduce a different form of complexity, which we call \emph{stance entanglement}. As shown in Fig.~\ref{fig:intro} (Right), these tasks are complex because they require simultaneous reasoning from the stances of all stakeholders. The decision of each stakeholder depends on those of others, which in turn depend on it. Thus, different stances are coupled through a mutual-dependence loop and become entangled within a single reasoning trajectory. As the number of stakeholders involved increases, reasoning over entangled stances exceeds the capability of a single LLM call~\cite{wu2023hitom,kim2023fantom}. 
More importantly, these mutually dependent stances cannot be solved in isolation, resisting standard divide-and-conquer MAS solutions. This motivates us to design a new MAS paradigm for stance disentanglement.

To achieve this, we begin by examining what constitutes a good decision. At its core, a good decision maximizes payoff while exhibiting no exploitable weaknesses, meaning that it lies in an equilibrium from which no stakeholder can improve payoff through unilateral deviation. Otherwise, the current decision is suboptimal and either can deviate to obtain a higher payoff or can be exploited when others deviate. This characterization parallels the definition of a Nash equilibrium~\cite{nash1951non} and motivates us to draw inspiration from how equilibria are solved in game theory.
Among methods for equilibrium solving~\cite{Lemke1964EquilibriumPO, Koller1994FastAF, Sandholm2005MixedIntegerPM}, fictitious play~\cite{brown1951fictitious-play} offers a solution to disentangle coupled strategic stances: it converts the mutually dependent fixed point solving problem into an iteratively convergent process in which each player simply best responds to the empirical average of others' past strategies. Moreover, its mechanism is well-suited to LLM-based agents, as it reduces each step to a single-chain prediction that an LLM can naturally produce~\cite{kim2023fantom, mccoy2024language}, providing the game-theoretic foundation for our MAS design.

Building on this insight, we propose the Multi-Agent Fictitious Play (MAFP) algorithm, which leverages MAS as a simulator and derives decisions through multi-agent co-evolution. Specifically, we decompose stances into agents, each representing a stakeholder. The algorithm proceeds in iterative rounds. Within each round, agents update their decisions by best responding to the empirical mixture of others' past decisions. By doing so, agents probe one another's exploitable weaknesses and co-evolve to reduce exploitability and improve payoffs. 
To realize this in natural-language space, MAFP introduces two operators: an \emph{aggregation operator} that constructs the empirical mixture of a set of decisions, and a \emph{best-response operator} that generates a new decision that maximizes utility against others' decisions. After the final round, the empirical mixture serves as the framework's output.

We evaluate MAFP on decision-making tasks across 13 scenarios, spanning competitive games \cite{duan2024gtbench} and negotiation \cite{bianchi2024negoarena}. These tasks require the model to decide the strategy for each scenario before acting by generating an open-ended language policy, a challenging problem due to the vast space of possible strategies. We report two complementary metrics: tournament strength, which measures a method's average payoff against all candidates being evaluated, and robustness, which captures its worst-case performance against adversaries that actively adapt to exploit it. Experimental results show that MAFP outperforms both single-round and multi-round baselines on both metrics, validating it as a promising MAS framework for decision-making.
Our contributions are summarized as follows:
\begin{itemize}[leftmargin=*]
    \item We identify \emph{stance entanglement} as a new form of complexity lies in real-world decision-making tasks that poses new challenge to the \emph{execution complexity} addressed by existing MAS.
    \item Inspired by \emph{fictitious play} from game theory, we propose \emph{Multi-Agent Fictitious Play (MAFP)}, a multi-agent framework that decomposes entangled stances to agents and leverages MAS as a simulator for robust decision finding.
    \item We propose a dual-axis evaluation method for good decisions along two complementary axes: \emph{tournament strength} and \emph{robustness}. Experiments across 13 scenarios demonstrate MAFP's effectiveness.
\end{itemize}

\section{Related Works}
In this section, we review recent research related to MAFP from two perspectives: how existing LLM-based MAS work and how decision-making tasks are addressed.

\subsection{LLM-based Multi-Agent System}

MAS~\cite{liu2024dynamic, li2023camel} demonstrates effectiveness on a wide range of complex tasks~\cite{xu2025deepresearch, deng2025swe-pro, Merrill2026TerminalBenchBA, Wei2025BrowseCompAS} by having multiple LLM-based agents work collaboratively. These systems target tasks with execution complexity and operate through a divide-and-conquer paradigm. Tasks that demand long execution chains, such as software engineering~\cite{deng2025swe-pro} and deep research~\cite{xu2025deepresearch}, are decomposed across role-specialized agents and executed sequentially~\cite{qian2024chatdev, hong2024metagpt}; those that demand broad information coverage or divergent exploration, such as wide research~\cite{manus2025wideresearch} and multi-agent debate~\cite{liang2024encouraging}, are executed in parallel~\cite{du2024improving, wang2025mixture}.

Existing research primarily operates within this cooperative paradigm, focusing on how to split and arrange tasks effectively, spanning orchestrator optimization~\cite{wu2024autogen, chen2024agentverse, team2026kimi} and topology optimization~\cite{zhuge2024gptswarm, yang2026agentnet, xu2026tacomas}. 
However, all of these remain within a split-and-aggregate formulation, and fall short on decision-making tasks~\cite{duan2024gtbench, chen2023aucarena, backlund2025vending} with interdependent stances. 
Even multi-agent debate, the line that most directly surfaces competing views, only aggregates independently authored solutions without reducing the reasoning complexity introduced by stance entanglement.
In this paper, we address such tasks through multi-agent fictitious play, a new MAS paradigm that decomposes stances into self-interested agents and treats MAS as a simulator for decision finding.

\subsection{LLM for Decision Making}

As LLMs~\cite{yang2025qwen3, singh2025gpt5, liu2026mtp} grow more capable, they are increasingly deployed as strategic decision-makers, and a wave of benchmarks evaluates this capability across games~\cite{duan2024gtbench,zhang2024mastermind,costarelli2024gamebench,akata2025playing,chen2024llmarena}, negotiation~\cite{bianchi2024negoarena, meta2022human, rivera2024escalation}, social deduction~\cite{guo2023suspicion, bailis2024werewolf, light2023avalonbench}, and competitive market~\cite{chen2023aucarena, backlund2025vending, wang2025bits, yun2025quantevolve}. Existing attempts improve the decision-making capability of LLM through strategic reasoning, performing explicit theory-of-mind~\cite{frith2005theory} (ToM) reasoning at inference time.
Their solutions range from one-step perspective-taking on opponents~\cite{wilf2024think, cross2024hypothetical}, to two-order reasoning that additionally anticipates how others perceive the agent~\cite{wang2024boosting}, to recursive level-$k$ mutual anticipation that extends ToM to arbitrary depth~\cite{zhang2025klevel}. What unites these methods is the assumption that an LLM can carry out higher-order belief reasoning within a single inference pass.

However, recursive mutual anticipation is a structural weak point of LLMs, which are essentially autoregressive next-token predictors~\cite{mccoy2024embers, mccoy2024language}. This limitation is verified empirically by Hi-ToM~\cite{wu2023hitom} and FANToM~\cite{kim2023fantom}, 
showing that LLM accuracy drops rapidly in higher-order belief reasoning (e.g., ``I think that you think'') and exhibits an illusory theory of mind. As the number of stakeholders involved increases in real-world tasks, the breadth and depth of ToM reasoning exceed the capability of a single LLM call.
In this work, we propose an MAS solution for decision-making that decomposes stances into agents and derives decisions through multi-agent co-evolution.
By doing so, each LLM call reduces to a single-layer logic reasoning, aligning with LLMs' strengths in conditional prediction.

\section{MAFP: Multi-Agent Fictitious Play for Robust Decision Making}

We propose Multi-Agent Fictitious Play (MAFP), a training-free framework for robust decision making over natural-language policies (Section~\ref{sec:problem}). 
Inspired by fictitious play~\cite{brown1951fictitious-play,monderer1996fictitious}, MAFP addresses the mutual-anticipation dilemma (Section~\ref{sec:dilemma}) with multi-agent co-evolution, in which agents iteratively best respond to each other's policies and co-evolve toward robust profiles through repeated rounds of update (Section~\ref{sec:algo}).
We elaborate on the design of MAFP in the following.

\subsection{Problem Formulation}\label{sec:problem}

\paragraph{Decision Making Task.}
We consider strategic decision-making expressed as a language-policy game. A scenario is specified by a natural-language description $\mathcal{D}$ together with a set of stakeholders $\mathcal{N}=\{1,\ldots,n\}$, where $n$ is the number of stakeholders.

Each stakeholder \zy{$i$} is characterized by a \emph{stance} that summarizes its situational profile:
\begin{equation}
\omega_i \;=\; (r_i,\, g_i,\, c_i,\, \rho_i),
\end{equation}
\zy{where the stance $\omega_i$ comprises its role $r_i$, goal $g_i$, private context and constraints $c_i$, and payoff description $\rho_i$.}
Each stakeholder \zy{$i$ then} commits to a textual strategy describing how to take action, which we call a language policy, denoted by $\pi_i$:
\begin{equation}
\pi_i \;\in\; \Pi_i,
\end{equation}
where $\Pi_i$ is the space of natural-language strategies admissible for stakeholder $i$. The joint commitment of all stakeholders yields a \emph{policy profile}, denoted by $\boldsymbol{\pi}$:
\begin{equation}
\boldsymbol{\pi} \;=\; (\pi_1,\ldots,\pi_n) \;\in\; \Pi
\;=\; \prod_{i=1}^{n}\Pi_i,
\end{equation}
where $\Pi$ is the joint policy space.

Since the utility of a policy profile $\boldsymbol{\pi}$ inherently depends on the opponents it plays against, we evaluate it in expectation over a reference distribution of opponent profiles. We thus define the utility $U_\mathcal{P}(\boldsymbol{\pi})$ as the expected \zy{payoff} of $\boldsymbol{\pi}$ against profiles drawn from a given distribution $\mathcal{P}$.
To estimate the \zy{utility}, each stakeholder's language policy conditions an identical action model $M_{\mathrm{act}}$ that produces actions step by step:
\begin{equation}\label{eq:action-model}
a_{i,t} \;\sim\; M_{\mathrm{act}}\!\left(\,\cdot \mid s_t, \mathcal{D},
\omega_i, \pi_i\,\right),
\end{equation}
where $a_{i,t}$ denotes the action taken by stakeholder $i$ at time step $t$ from state $s_t$.
The environment manages state transitions according to its own rule-based dynamics, and a full match yields a trajectory $\tau$:
\begin{equation}
\tau = (s_0, a_0, s_1, a_1, \ldots, s_T)  \;\sim\; 
\mathrm{Env}\!\left(\mathcal{D};\, \boldsymbol{\pi}\right) 
\end{equation}
where $\mathrm{Env}(\cdot)$ denotes the environment's transition dynamics, 
\zy{with each $\pi_i$ driving stakeholder $i$'s actions through $M_{\mathrm{act}}$.} 
From this trajectory, the environment computes a rule-based payoff $R_i(\tau)$ for each stakeholder $i$.

The per-stakeholder \zy{utility} of policy $\pi_i$ against the distribution $\mathcal{P}$ is then defined as
\begin{equation}\label{eq:per-utility}
U_{\mathcal{P}}(\pi_i) \;=\;
\mathbb{E}_{\hat{\boldsymbol{\pi}} \sim \mathcal{P}} U_{(\pi_{i},\hat{\boldsymbol{\pi}}_{-i})}(\pi_i) \;= \;
\mathbb{E}_{\hat{\boldsymbol{\pi}} \sim \mathcal{P}}\;
\mathbb{E}_{\tau \sim \mathrm{Env}(\mathcal{D};\, \pi_i, \hat{\boldsymbol{\pi}}_{-i})}
\!\left[R_i(\tau)\right],
\end{equation}
\zy{where $\hat{\boldsymbol{\pi}}$ is a profile sampled from $\mathcal{P}$, $\hat{\boldsymbol{\pi}}_{-i}$ denotes its components corresponding to stakeholders other than $i$, and $(\pi_i, \hat{\boldsymbol{\pi}}_{-i})$ is the joint profile in which $i$ plays $\pi_i$ while the remaining roles are filled by $\hat{\boldsymbol{\pi}}_{-i}$.}
The overall utility of the policy profile $\boldsymbol{\pi}$ against $\mathcal{P}$ is the average of its per-stakeholder \zy{utilities}:
\begin{equation}\label{eq:utility}
U_{\mathcal{P}}(\boldsymbol{\pi}) \;=\;
\frac{1}{n} \sum_{i=1}^{n} U_{\mathcal{P}}(\pi_i).
\end{equation}

\paragraph{Finding Robust Policy is Equilibrium Seeking.}

In strategic environments, the other stakeholders are themselves goal-driven, and the profile $\boldsymbol{\pi}_{-i}$ that stakeholder $i$ faces is not fixed: once others observe and infer $i$'s policy, they adjust toward whatever better serves their own goals. A good policy therefore cannot be obtained by simply maximizing one's own utility alone; it must also account for the other stakeholders' goals, staying
advantageous even after they deviate toward what better serves them. We formalize this requirement through the notion of \emph{unilateral deviation}. For a given profile $\boldsymbol{\pi}$, the deviation gain available to stakeholder $i$, denoted by $\Delta_i(\boldsymbol{\pi})$, is the utility improvement it can attain by switching to its best alternative policy while the others remain fixed:
\begin{equation}
\Delta_i(\boldsymbol{\pi}) \;=\;
\max_{\pi_i'\in\Pi_i}
U_{(\pi_{i}',\boldsymbol{\pi}_{-i})}(\pi_i')
\;-\;
\zy{U_{(\pi_{i},\boldsymbol{\pi}_{-i})}(\pi_i)},
\end{equation}
where $\pi_i'$ ranges over alternative language policies in $\Pi_i$, and 
$U_{(\pi_{i}', \boldsymbol{\pi}_{-i})}(\pi_{i}') = \mathbb{E}_{\tau \sim \mathrm{Env}(\mathcal{D};\, \pi_{i}', {\boldsymbol{\pi}}_{-i})}[R_i(\tau)]$ denotes the utility of stakeholder $i$ under the joint profile $(\pi_i', \boldsymbol{\pi}_{-i})$; the term \zy{$U_{(\pi_i, \boldsymbol{\pi}_{-i})}(\pi_i)$} is defined analogously.

Accordingly, finding a good decision amounts to reaching a profile at which no stakeholder retains any incentive to deviate, i.e., the largest deviation gain equals zero:
\begin{equation}\label{eq:no-deviation}
\max_{i \in \mathcal{N}} \Delta_i(\boldsymbol{\pi}) \;=\; 0 .
\end{equation}
If $\Delta_i(\boldsymbol{\pi}) > 0$ for some stakeholder $i$, then $i$ can improve its utility by deviating to a better policy $\pi_i'$. More importantly, such a deviation propagates across the profile: under the new profile  $(\pi_i', \boldsymbol{\pi}_{-i})$, the other stakeholders' utilities are altered, and their deviation gains $\Delta_j$ may also shift. 
For example, after $i$ switches from $\pi_i$ to $\pi_i'$, some other stakeholder $j$ may see its utility drop while $\Delta_j$ simultaneously grows, revealing that $\pi_j$ is itself far from optimal and likewise admits further improvement. 
This mutual interdependence implies that any profile with $\max_{i \in \mathcal{N}} \Delta_i(\boldsymbol{\pi}) > 0$ is inherently sub-optimal.

Conversely, when $\max_{i \in \mathcal{N}} \Delta_i(\boldsymbol{\pi}) = 0$, no stakeholder can gain by a unilateral deviation; equivalently, $\boldsymbol{\pi}$ is a Nash equilibrium~\cite{nash1951non} of the game~\cite{lockhart2019computing}.
Therefore, the task of robust decision making is fundamentally an \emph{equilibrium-seeking} process.

\begin{figure}
    \centering
    \includegraphics[width=1.0\linewidth]{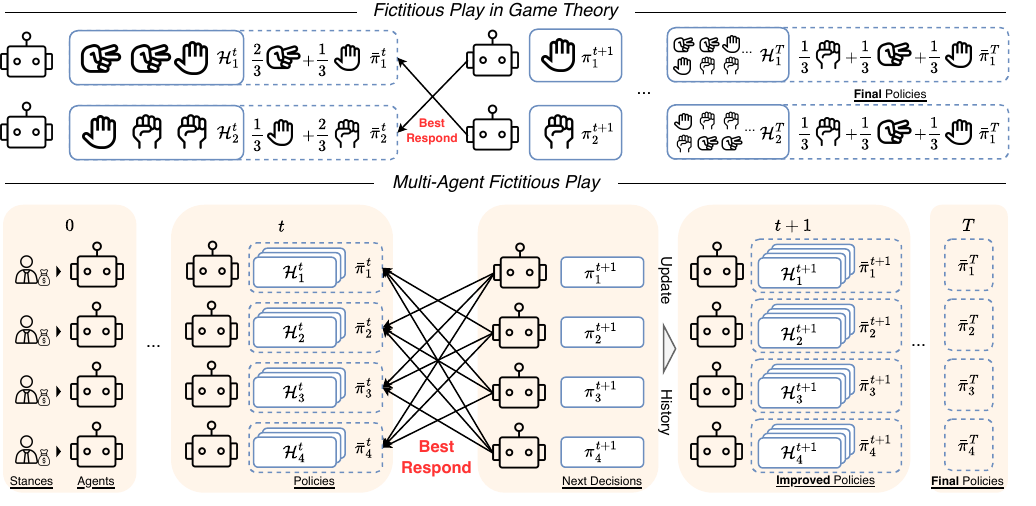}
    \caption{\textbf{Illustration of MAFP algorithm.} \emph{Fictitious play in game theory} finds equilibrium through an iteratively convergent process in which each player best responds to the empirical average of others’ past actions, here converging to the Nash equilibrium of rock–paper–scissors. Inspired by this, \emph{multi-agent fictitious play (MAFP)} decomposes stances into agents and finds policies through multi-agent co-evolution: at each round, agents update decisions by best-responding to the empirical mixture of others’ past decisions.}
    \label{fig:algo}
\end{figure}
\subsection{From Mutual Anticipation to Fictitious Play}\label{sec:dilemma}

\paragraph{Mutual Anticipation and Its Recursive Dilemma.}
Solving an equilibrium is difficult for LLMs because each stakeholder's optimal policy depends on the others' policies, which in turn depend on their expectations about us. This mutual anticipation results in a search space explosion: if $b_i$ denotes the branching factor for each stakeholder and $d$ is the depth of the belief hierarchy, the reasoning tree grows exponentially as
\begin{equation}
\Big(\prod_{i=1}^{n} b_i\Big)^{d}.
\end{equation}
Even modest depths make the search intractable, as the policy spaces of language-policy games are vast. We refer this as \emph{recursive mutual-anticipation dilemma}.

\paragraph{Fictitious Play as a Decomposition of the Recursion.}
\textit{Fictitious play}~\cite{brown1951fictitious-play, monderer1996fictitious} in \zy{{game theory}} sidesteps this difficulty by spreading the recursion across discrete iterations. As illustrated in Fig.~\ref{alg:MAFP}, instead of unrolling the belief hierarchy inside a single deliberation, each agent at round $t$ best-responds to the empirical average of opponents' past policies:
\begin{equation}\label{eq:fp}
\pi_i^{\,t+1} \;\in\; \arg\max_{\pi_i \in \Pi_i}
U_{(\pi_i,\,\boldsymbol{\bar{\pi}}_{-i}^{\,t})}(\pi_i),
\qquad
\boldsymbol{\bar{\pi}}_{-i}^{\,t} \;=\; (\bar{\pi}_j^{\,t})_{j \neq i}
,
\qquad
\bar{\pi}_j^{\,t} \;=\; \mathrm{Avg}(\pi_j^{0},\ldots,\pi_j^{\,t}).
\end{equation}
where $\bar{\pi}_j^{\,t}$ denotes the empirical average of opponent $j$'s historical policies, and $\boldsymbol{\bar{\pi}}_{-i}^{\,t}$ collects these per-opponent averages into a joint profile.
By doing so, the recursive reasoning is replaced by iterative updates. As the iterations proceed, each opponent's empirical average policy captures how the opponent has adapted to the agent’s previous strategies, so responding to it implicitly accounts for multiple levels of anticipation.

This decomposition aligns well with LLMs' strengths, as each step is a flat reasoning task grounded in observed history. By transforming mutual anticipation into a sequence of best-response updates, fictitious play shifts the problem from ``reasoning \(K\) levels deep'' to ``reasoning one level deep, \(K\) times''.

\subsection{Textual Fictitious Play}\label{sec:algo}

Based on the principles described above, we now introduce the MAFP algorithm. As illustrated in Fig.~\ref{fig:algo}, MAFP operates entirely in language space through an LLM $M$: starting from a multi-agent initialization, each round applies an aggregation operator to obtain empirical mixtures and a best-response operator to update policies. After $K$ rounds, MAFP outputs a policy profile that specifies a policy for each stakeholder.

\paragraph{Multi-Agent Initialization.}
Given a decision-making task, MAFP instantiates a multi-agent system in which each agent represents a stakeholder $i \in \mathcal{N}$ for fictitious play. Each agent is initialized with a policy $\pi_i^0$ and a history set $\mathcal{H}$. The first policy is generated from the scenario description $\mathcal{D}$ and the stakeholder’s stance $\omega_i$,
\begin{equation}
\pi_i^{0} = M_{\mathrm{init}}(\mathcal{D}, \omega_i),
\qquad
\mathcal{H}_i^{0} = \{\pi_i^{0}\}.
\end{equation}

\paragraph{Aggregation Operator.}
At iteration $t$, each agent best responds based on its belief about how opponents are playing, which is the empirical average of their history (Eq.~\ref{eq:fp}). In language space, we obtain this belief using an aggregation operator $\mathrm{Agg}_M$ that prompts LLM to summarize historical policies
\begin{equation}
\bar{\pi}_j^{t} = \mathrm{Agg}_M\!\left(\mathcal{H}_j^{t}\right),
\qquad \forall\, j\in\mathcal{N}.
\end{equation}

\paragraph{Best-Response Operator.}
Then, agent $i$ produces a best-response on the obtained empirical-mixture opponent policy set $\bar{\boldsymbol{\pi}}_{-i}^{\,t} \;=\; (\bar{\pi}_j^{\,t})_{j \neq i}$.
This is realized through best-response operator $\mathrm{BR}_M$ conditioned on the scenario $\mathcal{D}$, the agent's own stance $\omega_i$, and the aggregated opponents $\bar{\boldsymbol{\pi}}_{-i}^{t}$,
\begin{equation}
\pi_i^{t+1} = \mathrm{BR}_M\!\left(\mathcal{D},\, \omega_i,\, \bar{\boldsymbol{\pi}}_{-i}^{t}\right),
\end{equation}
and the result is appended to the history, $\mathcal{H}_i^{t+1} = \mathcal{H}_i^{t} \cup \{\pi_i^{t+1}\}$.

\paragraph{Final Per-stakeholder Policy.}
After $K$ rounds, we take the empirical mixture of all policies generated by each agent as its final policy, which captures the adaptations it accumulated in response to every weakness that other agents exploited across the rounds. Concretely, we use the same aggregation operator $\mathrm{Agg}_M$ to aggregate each stakeholder's history into one executable policy:
\begin{equation}
\pi_i^{\mathrm{out}} = \mathrm{Agg}_M\!\left(\mathcal{H}_i^{K}\right),
\end{equation}
which is the final policy for stance $i$. MAFP returns the policy profile with policies for all stances $\boldsymbol{\pi}^{\mathrm{out}} = (\pi_1^{\mathrm{out}},\, \ldots,\, \pi_n^{\mathrm{out}})$.
The full procedure is presented in Algorithm~\ref{alg:MAFP}.

\begin{algorithm}[t]
\caption{Multi-Agent Fictitious Play over Language Policies}
\label{alg:MAFP}
\begin{algorithmic}[1]
\REQUIRE Scenario $\mathcal{D}$, stances $\{\omega_i\}_{i=1}^{n}$, frozen LLM $M$, iterations $K$
\ENSURE Final policy profile $\boldsymbol{\pi}^{\mathrm{out}}$

\FOR{each stakeholder $i\in\mathcal{N}$}
    \STATE Initialize $\pi_i^0 = M_{\mathrm{init}}(\mathcal{D},\omega_i)$ and $\mathcal{H}_i^0=\{\pi_i^0\}$
\ENDFOR

\FOR{$t=0$ to $K-1$}
    \FOR{each stakeholder $j\in\mathcal{N}$ \textbf{in parallel}}
        \STATE Aggregate empirical mixture:\quad
        $\bar{\pi}_j^{t} = \mathrm{Agg}_M\!\left(\mathcal{H}_j^{t}\right)$
    \ENDFOR
    \FOR{each stakeholder $i\in\mathcal{N}$ \textbf{in parallel}}
        \STATE Best response against $\bar{\boldsymbol{\pi}}_{-i}^{t}=\{\bar{\pi}_j^{t}\}_{j\neq i}$:\quad
        $\pi_i^{t+1} = \widehat{\mathrm{BR}}_M\!\left(\mathcal{D},\omega_i,\bar{\boldsymbol{\pi}}_{-i}^{t}\right)$
        \STATE Update history:\quad $\mathcal{H}_i^{t+1}=\mathcal{H}_i^t\cup\{\pi_i^{t+1}\}$
    \ENDFOR
\ENDFOR

\FOR{each stakeholder $i\in\mathcal{N}$}
    \STATE $\pi_i^{\mathrm{out}} = \mathrm{Agg}_M\!\left(\mathcal{H}_i^{K}\right)$
\ENDFOR

\RETURN $\boldsymbol{\pi}^{\mathrm{out}}=(\pi_1^{\mathrm{out}},\ldots,\pi_n^{\mathrm{out}})$
\end{algorithmic}
\end{algorithm}

\section{Experiments}

In this section, we conduct experiments to investigate two research questions: 
\textbf{RQ1:} How does fictitious play improve the performance compared to existing test-time scaling frameworks?
\textbf{RQ2:} How does utility improve as the number of rounds increases?

\subsection{Experimental Setup}
\label{sec:exp-setup}

\paragraph{Policy Generation Benchmark.}
We evaluate \textsc{MAFP} on 13 scenarios spanning two strategic
decision-making categories: competitive games~\cite{duan2024gtbench} and
natural-language negotiation~\cite{bianchi2024negoarena}, representing strategic games with clear rules and language-based real-world tasks. The selected scenarios exhibit diversity in characteristics across dimensions, such as whether it is a zero-sum game, static or dynamic, and whether information is complete, to comprehensively reflect the algorithm's features. Detailed scenario description can be found in Appendix~\ref{app:scenarios}. To isolate the differences between policy profiles, all of which are executed by Qwen3.5-35B-A3B~\cite{yang2025qwen3} as the action model $\mathcal{M}_{\text{act}}$
(Eq.~\ref{eq:action-model}). To ensure the reliability of the results, we play $16$ matches with seat exchange for each pair, averaging the payoff of each scenario. And the full pipeline is run with 8 random seeds and we report seed-averaged means. 

\paragraph{Metrics.}
We evaluate each generated policy profile $\boldsymbol{\pi}$ along two complementary axes, \emph{Tournament Strength} and \emph{Robustness}, both measured by the utility $U_\mathcal{P}(\boldsymbol{\pi})$ of Eq.~\ref{eq:utility} but instantiated with different opponent distributions $\mathcal{P}$. 
Notably, the two axes correspond to the two ways a policy can violate the equilibrium criterion of Eq.~\ref{eq:no-deviation}: either a higher-utility deviation still exists for it (so it is not strong enough), or it leaves a weakness that an adapting opponent can exploit (so it is not robust enough). A good policy must rule out both.

\emph{Tournament Strength} (TS) instantiates Eq.~\ref{eq:utility} with
$\mathcal{P}\!=\!\mathcal{P}_{\text{cand}}$, the empirical distribution over the other competing profiles $\mathcal{P}_{\text{cand}}$ in the experiment. TS is estimated using a round-robin schedule, in which profiles engage in pairwise battles. A policy that is beaten by the field admits a stronger response and thus retains a profitable deviation; higher TS therefore means the policy is closer to optimal against the current candidates.

\emph{Robustness} (Rob) measures the extent to which a target policy retains its utility when other players adapt against it: if an adapting opponent can drive the utility down, the policy leaves exploitable weaknesses.
Concretely, we calculate the robustness of a profile $\mathrm{Rob}(\boldsymbol{\pi})$ by averaging the robustness of each policy within it
\begin{equation}
    \mathrm{Rob}(\boldsymbol{\pi}) \coloneqq \frac1n\sum_{i=1}^n \mathrm{Rob}(\pi_i).
\end{equation}
For each policy $\pi_i$, freeze it and let an attacker LLM evolve a counter-profile
$\boldsymbol{\pi}_{-i}^{(R)}$ over $R\!=\!10$ rounds, where each round plays $4$ matches against $\pi_i$ to gather enough evidence. The evolution is achieved by rewriting the policies within the counter-profile based on the collected evidence to maximize their respective utility. $\mathrm{Rob}(\pi_i)$ is then $\pi_i$'s lowest utility against the evolved profile across $R$ rounds,
\begin{equation}
\mathrm{Rob}(\pi_i) \coloneqq \min_{r \in [R]}\; U_{(\pi_i,\boldsymbol{\pi}_{-i}^{(r)})}(\pi_i).
\end{equation}

\subsection{RQ1: MAFP versus Existing Test-Time Scaling Frameworks}
\label{sec:exp-main}

\begin{table}[t]
\centering
\caption{\textbf{Comparison Results on Tournament Strength.} For each scenario, all 9 methods are evaluated by a round-robin tournament, measuring each method's average utility against the other 8. ``Avg.'' denotes the mean utility across the 13 scenarios. Best per column in bold.}
\label{tab:p2_wr_avg_seat_T}
\setlength{\tabcolsep}{3pt}
\resizebox{\textwidth}{!}{%
\begin{tabular}{l|ccccccccccccc|c}
\toprule
Method & TicTac & Nim & IPD & Conn4 & Pig & BrkThru & KuhnPk & BlindAuc & LiarDice & Negot & BuySell & Ultimat & ResExch & Avg. \\
\midrule
\multicolumn{15}{c}{\textit{Single Round}} \\
\midrule
Q3-1.7B        & 0.616 & \textbf{0.624} & 0.548 & 0.509 & 0.357 & \textbf{0.734} & 0.505 & 0.476 & 0.243 & 0.600 & 0.361 & 0.382 & 0.496 & 0.496 \\
Llama-3.1-8B   & 0.490 & 0.456 & 0.526 & 0.520 & 0.508 & 0.458 & 0.459 & 0.399 & 0.258 & \textbf{0.614} & 0.331 & 0.365 & 0.499 & 0.452 \\
GPT-5-nano     & 0.383 & 0.342 & 0.443 & 0.421 & 0.382 & 0.390 & 0.406 & 0.367 & 0.351 & 0.590 & 0.429 & 0.466 & 0.494 & 0.420 \\
Q3.5-35B       & 0.458 & 0.601 & 0.386 & \textbf{0.559} & 0.493 & 0.549 & 0.493 & 0.384 & 0.547 & 0.593 & 0.413 & 0.535 & 0.495 & 0.500 \\
\midrule
\multicolumn{15}{c}{\textit{Multiple Round with Q3.5-35B}} \\
\midrule
SR       & 0.544 & 0.458 & 0.347 & 0.491 & 0.455 & 0.461 & 0.477 & 0.362 & 0.544 & 0.599 & 0.505 & 0.464 & 0.497 & 0.477 \\
Debate   & 0.468 & 0.578 & 0.385 & 0.484 & 0.535 & 0.513 & 0.441 & 0.506 & 0.490 & 0.505 & 0.552 & 0.486 & 0.502 & 0.496 \\
ToM      & \textbf{0.624} & 0.494 & 0.385 & 0.521 & 0.479 & 0.575 & 0.534 & 0.611 & 0.550 & 0.452 & \textbf{0.573} & \textbf{0.559} & 0.499 & 0.527 \\
\midrule
MAFP-Last  & 0.477 & 0.375 & \textbf{0.899} & 0.428 & 0.469 & 0.420 & 0.505 & \textbf{0.630} & 0.373 & 0.299 & 0.553 & 0.459 & 0.498 & 0.491 \\
MAFP       & 0.531 & 0.485 & 0.500 & 0.537 & \textbf{0.557} & 0.542 & \textbf{0.583} & 0.604 & \textbf{0.551} & 0.504 & 0.472 & 0.549 & \textbf{0.508} & \textbf{0.533} \\
\bottomrule
\end{tabular}%
}
\end{table}

\begin{table}[t]
\centering
\caption{\textbf{Comparison Results on Robustness.} For each scenario, methods are evaluated by freezing each as the target and letting an attacker evolve a counter-profile over 10 rounds; robustness is the target's lowest utility across those rounds. ``Avg.'' denotes the mean robustness across the 13 scenarios. Best per column in bold.}
\label{tab:p2_rob_min_seat_T}
\setlength{\tabcolsep}{3pt}
\resizebox{\textwidth}{!}{%
\begin{tabular}{l|ccccccccccccc|c}
\toprule
Method & TicTac & Nim & IPD & Conn4 & Pig & BrkThru & KuhnPk & BlindAuc & LiarDice & Negot & BuySell & Ultimat & ResExch & Avg. \\
\midrule
\multicolumn{15}{c}{\textit{Single Round}} \\
\midrule
Q3-1.7B        & 0.232 & \textbf{0.477} & 0.133 & 0.406 & 0.174 & \textbf{0.531} & 0.305 & 0.348 & 0.363 & \textbf{0.496} & 0.234 & 0.473 & 0.500 & 0.359 \\
Llama-3.1-8B   & 0.227 & 0.312 & 0.018 & \textbf{0.434} & 0.281 & 0.203 & 0.246 & 0.328 & 0.402 & 0.467 & 0.334 & 0.354 & 0.500 & 0.316 \\
GPT-5-nano     & 0.164 & 0.078 & 0.031 & 0.320 & 0.281 & 0.312 & 0.219 & 0.234 & 0.430 & 0.443 & 0.447 & 0.562 & 0.500 & 0.309 \\
Q3.5-35B       & 0.258 & 0.297 & 0.031 & 0.406 & 0.340 & 0.266 & 0.277 & 0.328 & 0.531 & 0.406 & 0.385 & 0.584 & 0.500 & 0.355 \\
\midrule
\multicolumn{15}{c}{\textit{Multiple Round with Q3.5-35B}} \\
\midrule
SR       & 0.328 & 0.395 & 0.000 & 0.359 & 0.332 & 0.234 & 0.273 & 0.227 & 0.484 & 0.438 & 0.436 & 0.516 & 0.498 & 0.348 \\
Debate   & 0.229 & 0.328 & 0.023 & \textbf{0.434} & 0.258 & 0.297 & 0.266 & 0.367 & 0.516 & 0.398 & \textbf{0.559} & 0.549 & 0.502 & 0.363 \\
ToM      & 0.307 & 0.293 & 0.000 & 0.391 & 0.281 & 0.359 & \textbf{0.328} & 0.406 & 0.516 & 0.293 & 0.521 & 0.602 & 0.504 & 0.369 \\
\midrule
MAFP-Last  & 0.287 & 0.344 & \textbf{0.469} & 0.297 & 0.268 & 0.234 & 0.305 & \textbf{0.500} & 0.477 & 0.148 & 0.527 & 0.502 & 0.500 & 0.374 \\
MAFP       & \textbf{0.336} & 0.340 & 0.125 & 0.410 & \textbf{0.449} & 0.461 & 0.324 & 0.469 & \textbf{0.578} & 0.393 & 0.477 & \textbf{0.605} & \textbf{0.508} & \textbf{0.421} \\
\bottomrule
\end{tabular}%
}
\end{table}

To evaluate the effectiveness of MAFP, we compare it against a representative set
of test-time scaling frameworks. For single-round baselines, we adopt CoT reasoning
with four different LLM backbones varying model families and capacity: Qwen3-1.7B, Llama-3.1-8B-Instruct~\cite{grattafiori2024llama3}, GPT-5-nano~\cite{singh2025gpt5}, and Qwen3.5-35B-A3B~\cite{qwen35blog}. 
For multi-round baselines, we take the strongest single-round backbone, Qwen3.5-35B-A3B, as the unified backbone for four multi-round frameworks: self-reflection (SR)~\cite{shinn2023reflexion}, debate~\cite{liang2024encouraging}, theory-of-mind~\cite{wilf2024think} (ToM), and our MAFP. Implementation details can be found in Appendix~\ref{app:baselines}.
To investigate the contribution of MAFP's aggregation step, we additionally design an MAFP-Last variant, an ablation that removes aggregation and best responds to the opponent's latest policy at each round rather than to the empirical mixture of past iterates.
All multi-round baselines use a unified 4-round iteration to ensure fair comparison.
Across these 9 methods, we calculate Tournament Strength and Robustness metrics, per-scenario
results are summarised in Tables~\ref{tab:p2_wr_avg_seat_T}~and~\ref{tab:p2_rob_min_seat_T}.

From the results, MAFP achieves the highest TS ($0.533$) and the highest Rob ($0.421$) on average against all other candidates. This validates our main claim that multi-agent co-evolution through fictitious play yields decisions that are both strong and robust.
Among the multi-round baselines, SR and Debate, despite introducing iterative updates, fail to achieve meaningful improvements over single-round baselines, confirming that iteration alone cannot resolve the recursive mutual-anticipation dilemma. ToM partially addresses this dilemma by incorporating explicit other-agent reasoning, lifting both metrics above all single-round and non-modeling multi-round baselines.
However, it still unfolds within a single chain and does not escape the recursive mutual-anticipation process, leaving it behind MAFP especially on robustness (0.369 vs. 0.421). 
These results validate MAFP's core design motivation: producing \emph{robust} policies by replacing single-chain recursive anticipation with iterative best responses to an empirical mixture of historical policies, which distributes anticipation across rounds rather than unfolds it with single-chain reasoning.

The aggregation-removed ablation, MAFP-Last, trails MAFP on both metrics. This confirms that best-responding to the empirical mixture instead of the latest policy drives performance gains, which aligns with the classical fictitious play algorithm. Without aggregation, MAFP-Last 's greedy reaction to the latest iterate captures only the most recent strategic shift and discards the multiple levels of anticipation inherent in the history.

While MAFP achieves the highest TS and Rob averages, it does not exhibit consistent dominance across all 13 scenarios. For example, on TS, MAFP falls short on deterministic games with perfect information, including TicTacToe, Nim, ConnectFour, and Breakthrough, where a strong strategy can be found by local, position-by-position 
backward induction~\cite{kuhn1953extensive}, which single-step CoT can already handle well. In contrast, MAFP excels in scenarios with imperfect information, stochastic transitions, or general-sum payoffs, such as Pig, Kuhn Poker, and Liar's Dice, where no dominant pure strategy exists and a robust policy must hedge against an entire distribution of other players' behaviors.
Another notable exception is the Iterated Prisoner’s Dilemma (IPD), where MAFP-Last beats MAFP by a large margin on both TS and Rob. This is because Prisoner’s Dilemma is dominance-solvable, where Defect strictly dominates Cooperate. As a result, the last-iterate best response quickly collapses to pure defection, an unexploitable Nash strategy.
Overall, MAFP excels in complex scenarios with imperfect information, stochastic transitions, or a non-trivial mixed equilibrium structure, suggesting greater potential for real-world tasks.

\subsection{RQ2: Convergence Behaviour Across Iterations}
\label{sec:rq2}

\paragraph{Dynamics in Policy Generation.}
\begin{figure}
    \centering
    \includegraphics[width=1.0\linewidth]{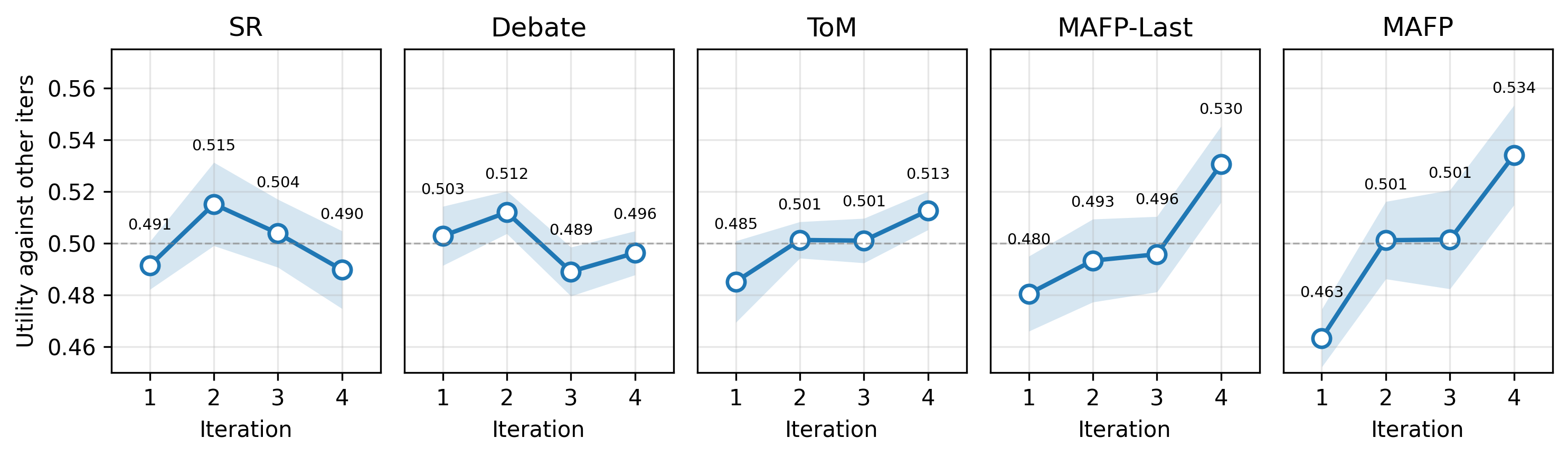}
    \vspace{-1.5em}
    \caption{\textbf{Per-iteration quality of policies produced by each
  iterative method.} For each method, we run an internal tournament among its four iterations and report each iteration's average utility against the other three. The shaded band shows the standard error of the mean.}
    \label{fig:rq2-policy}
\end{figure}

To answer RQ2, we examine the two iterative processes in our paper: the
\emph{policy-generation} process by which each method produces its final
profile, and the \emph{robustness-measurement} process by which an attacker
evolves a counter-profile against a frozen target. We conduct per-iteration
evaluations to visualize the dynamics of each.

Figure~\ref{fig:rq2-policy} reports the per-iteration quality of the five multi-round methods, computed by an internal round-robin tournament among the four policies produced at iterations 1–4 of the same method.  The results show that methods without other agent modeling, such as Debate and SR, struggle to benefit from additional iterations: they improve marginally in the first one or two rounds and then plateau or even regress in later rounds.
In contrast, methods that explicitly model other stakeholders---ToM, MAFP, and MAFP-Last---achieve meaningful gains across iterations, demonstrating the effectiveness of iterative refinement when policy updates are grounded in reasoning about other stakeholders. This contrast shows that naively switching from one-shot generation to iterative refinement is insufficient. Notably, MAFP and MAFP-Last achieve significantly larger improvements than ToM, validating their advantage in evolution.

\paragraph{Dynamics in Robustness Measurement.}
\begin{figure}
    \centering
    \includegraphics[width=1.0\linewidth]{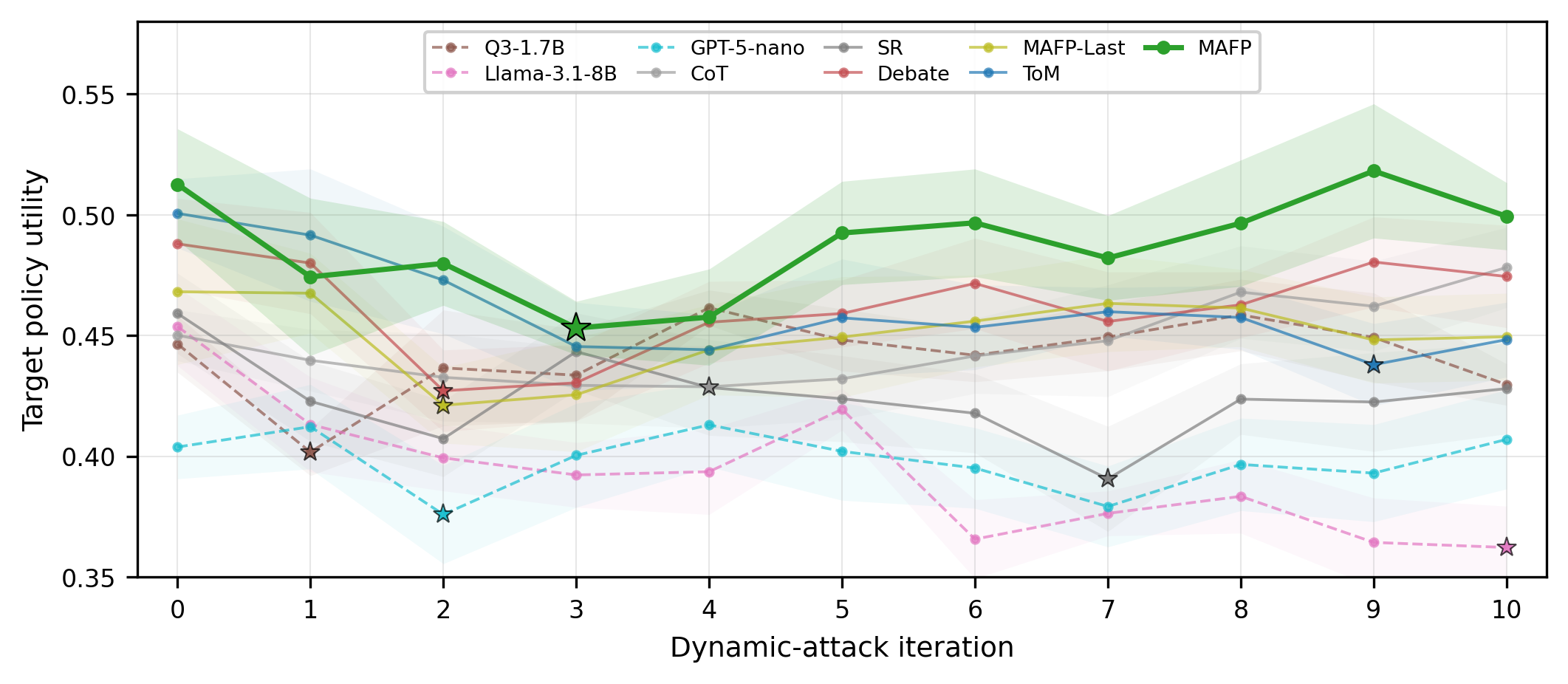}
    \vspace{-1.5em}
    \caption{\textbf{Target-profile utility under adversarial evolution
  during robustness evaluation.} Each curve shows a method's per-iteration utility against an evolving attacker, averaged across scenarios. The star marks each method's worst-case round. Shaded band shows the standard error of the mean.}
    \label{fig:rq2-robust}
\end{figure}
Figure~\ref{fig:rq2-robust} visualizes how the utility of the profile being evaluated evolves as the attacker is updated round-by-round
during robustness evaluation. Across all eight methods, utility decays in the early rounds, confirming our premise that an adaptive adversary can read the target's exposed behavior from past matches and rewrite its own policy to exploit it. 
For most methods, the lowest utility occurs in rounds~2-4. Subsequent evolution does not cause further declines and often increases the utility due to overfitting. This means our 10-round budget is enough for most methods to estimate the exploitability.

Crucially, MAFP's utility under exploitation exceeds that of other baselines at nearly every iteration.
This persistent gap substantiates our claim: the policy profile generated by MAFP is not merely strong on average but harder to exploit, which is the property that matters
most for decision making against strategic adversaries in the real world.

\section{Conclusion}\label{sec:conclusion}
In this work, we focus on enhancing the decision-making capability of LLMs through MAS. In contrast to the execution complexity that existing MAS are designed for, we identify stance entanglement as a different form of complexity introduced by decision-making. Drawing inspiration from fictitious play in game theory, we propose MAFP, a multi-agent framework that decomposes entangled stances to agents and leverages MAS as a simulator to derive the decision.
On a 13-scenario benchmark spanning competitive games and negotiation, MAFP attains the highest scores on both tournament strength and robustness among all evaluated baselines, with its advantage most pronounced in scenarios involving imperfect information, stochastic transitions, or mixed-strategy equilibria—precisely the conditions that characterize real-world strategic interaction.

\section{Limitations}
Two limitations point to natural extensions of MAFP. Experimentally, computational constraints confined our evaluation to the scenarios reported above. Our next step is to scale MAFP to richer real-world settings such as commercial decision-making in competitive market~\cite{backlund2025vending, yun2025quantevolve}. Such environments involve more stakeholders and more intricate strategic structure. We expect MAFP to demonstrate greater advantages through stances decomposition and fictitious play co-evolution.
Theoretically, MAFP rests on a clean game-theoretic formulation that opens room for deeper analysis: the convergence rate of language-space fictitious play to equilibrium, which equilibrium it selects when multiple Nash equilibria coexist~\cite{harsanyi1988general}, and how the iterative trajectory can be actively steered toward a desired equilibrium~\cite{zhang2023steering} are all high-impact open research problems we leave to future work.

{
\small
\bibliographystyle{plainnat}
\bibliography{ref}
}

\newpage
\appendix

\section{Implementation Details}

\subsection{Scenarios}\label{app:scenarios}
We evaluate MAFP across 13 scenarios, comprising 10 strategic games from GTBench~\citep{duan2024gtbench} and 3 negotiation settings from Negotiation Arena~\citep{bianchi2024negoarena}. These scenarios span a broad range of game-theoretic properties to comprehensively reflect the behavior of MAFP across different conditions. They include complete versus incomplete information, deterministic versus probabilistic dynamics, and zero-sum versus general-sum payoffs. We provide a per-scenario description of their rules and characteristics in Table~\ref{tab:scenarios}. 
Notably, our task is formulated differently from GTBench and Negotiation Arena, which treat the system being tested as the policy, prompting it to select an action at each step. Instead, we test the capability to generate policies: the tested system produces a natural-language policy specifying how to act before the game begins.

\begin{table}[t]
    \small
    \centering
    \caption{Scenarios Description.}
    \label{tab:scenarios}
    \begin{tabular}{lp{0.76\linewidth}}
        \toprule
        \textbf{Scenario} & \textbf{Description} \\
        \midrule
        \multicolumn{2}{c}{\textit{Strategic Games}} \\
        \midrule
        TicTacToe & A two-player game on a $3\times3$ grid; players alternate marking squares, and the first to align three marks horizontally, vertically, or diagonally wins. \\
        \midrule
        Connect~Four & Players alternately drop tokens into a $6\times7$ vertically suspended grid; the first to form a line of four wins. \\
        \midrule
        Breakthrough & An abstract strategy game played on a $3\times8$ board; pieces move one space straight or diagonally forward and may capture diagonally. The first player to reach the opponent's home row wins. \\
        \midrule
        Nim & Players alternately remove matches from one of four piles (initial sizes $1,3,5,7$); a player must remove at least one match from a single pile, and the player forced to take the last match loses. \\
        \midrule
        \makecell[l]{Iterated Prisoner's\\ Dilemma (IPD)} & Two players repeatedly choose between \texttt{Silent} and \texttt{Testify}; per-round payoffs follow the classic Prisoner's Dilemma matrix and accumulate over rounds. \\
        \midrule
        Pig & A turn-based dice game in which a player repeatedly rolls a single die, accumulating points until they choose to stop or roll a $1$ (losing the turn's gains); the first to reach the target score wins. \\
        \midrule
        Kuhn Poker & A two-player imperfect-information poker variant with a three-card deck (King, Queen, Jack); players alternately \texttt{Bet} or \texttt{Pass}, and the showdown awards the pot to the higher card. \\
        \midrule
        Blind Auction & Players simultaneously submit sealed bids for an item with private valuations; the higher bidder wins and pays their bid. \\
        \midrule
        Liar's Dice & A two-player game with private dice; players alternately bid increasing quantity-or-value combinations, or challenge the previous bid. The losing side of a challenge loses a die. \\
        \midrule
        Negotiation & Two players divide a pool of three item types with private value vectors, alternating between proposal turns and utterance turns; payoff is the total value of items each player ultimately receives. \\
        \midrule
        \multicolumn{2}{c}{\textit{Negotiations}} \\
        \midrule
        BuySell & A buyer with a private valuation and a seller with a private cost negotiate the price of a single item over multiple rounds; per-player surplus is the gap between the agreed price and the respective private value, with rejection zeroing both players. \\
        \midrule
        Ultimatum & A proposer offers a split of a fixed pot, and a responder either accepts (both receive the proposed shares) or rejects (both receive zero). \\
        \midrule
        Resource Exchange.& Two players hold different bundles of resources with asymmetric private values and negotiate trades over multiple rounds to maximize their own utility. \\
        \bottomrule
    \end{tabular}
\end{table}

\subsection{Payoff Definition}
All 13 scenarios are reduced to a single chess-style outcome 
$o\in\{0,0.5,1\}$ per match (seat-$0$'s score; seat-$1$'s is $1-o$), so that 
the reported win-rate is the simple mean of $o$ across matches. The 
reduction is identical for all scenarios within two structural groups, plus 
a universal illegal-move override.

Whenever a player emits a parse-failure or rule-illegal action, that player forfeits and the match is recorded as a loss for them, irrespective of any in-game scoring ($o=0$ if seat $0$ forfeits, $o=1$ if seat $1$ does). This rule dominates every other rule 
below.

The 10 strategic game scenarios 
each emit a canonical winner field $w\in\{P_0,P_1,\bot\}$, where $\bot$ 
denotes ``no winner'' (board filled without a line, equal cumulative 
scores, or turn-limit reached without resolution, depending on the game). 
The outcome is
\begin{equation}
o \;=\; \mathbb{1}[w=P_0] + \tfrac{1}{2}\,\mathbb{1}[w=\bot].
\end{equation}
The internal mechanism that produces $w$ varies between sub-groups but the 
conversion to $o$ does not. \emph{Board-completion} games (TicTacToe, 
Connect~Four, Breakthrough, Nim) decide $w$ from a terminal board pattern, 
with $\bot$ reachable when the board fills without one. 
\emph{Score-aggregation} games (IPD, Pig, Kuhn~Poker, Blind~Auction, 
Liar's~Dice, Negotiation) decide $w$ by comparing accumulated scores at 
game end---cumulative IPD payoff, points-to-target for Pig, pot size for 
Kuhn~Poker, value-minus-bid difference for Blind~Auction, last-with-chips 
for Liar's~Dice, and value totals from accepted deals for 
Negotiation---with $\bot$ reached when totals tie exactly.

For consistency across scenarios, we reduce the continuous payoffs of the 3 Negotiation Arena scenarios to discrete win/loss/draw outcomes by comparing the relative magnitudes of the two players' payoffs. Concretely, these scenarios each produce continuous per-player payoffs $(p_0,p_1)$, namely the value retained after the deal, with rejection zeroing both players. 
We collapse to the same discrete outcome via the sign-of-difference:
\begin{equation}
o \;=\; \mathbb{1}[p_0-p_1>\varepsilon] 
        + \tfrac{1}{2}\,\mathbb{1}[\lvert p_0-p_1\rvert\le\varepsilon],
\qquad \varepsilon = 10^{-6}.
\end{equation}

\subsection{Baselines}\label{app:baselines}

We compare MAFP against seven baselines drawn from two families: four single-round Chain-of-Thought policy authors that differ only in the authoring model, and three multi-round language-policy generators that share the same Qwen3.5-35B-A3B authoring backbone but differ in their frameworks.

\paragraph{Single-round baselines.}
Each baseline in this family follows the rules-only Chain-of-Thought prompt of \citet{wei2022cot}: the model receives the game's rules and a seat assignment (first or second player) and is asked to write a structured policy organized into labeled sections (Opening Principles, Midgame Priorities, Endgame / Closing Rules, Tactical Checks). One LLM call is issued per seat, and the two-seat outputs are bundled into a single per-scenario policy.

\paragraph{Multi-round baselines.}
The three multi-round baselines all use Qwen3.5-35B-A3B as the backbone and run for $K{=}4$ rounds, matching the
depth used by MAFP. They differ in what each round consumes as input:
\begin{itemize}
    \item \textbf{SR (Self-Reflection).} We follow
    \citet{shinn2023reflexion} to implement verbal self-critique at
    the policy level. At each round $k\geq 2$, each seat is shown
    only its own previous (round-$k{-}1$) policy and is asked to
    critique and refine it; no opponent policy and no game trace is
    introduced.

    \item \textbf{Debate.} We follow
    \citet{liang2024encouraging} to implement multi-agent debate. In round 1, $N=2$ author agents independently propose a policy profile from rules. In rounds $k=2,\dots, K$, each agent refines its profile after seeing the other agent's previous profile. After round $K$, a judge LLM calls aggregates the $N$ final profiles into a single consensus profile.

    \item \textbf{ToM ($K$-level reasoning).} We follow
    \citet{zhang2025klevel} to adapt their recursive $K$-level
    reasoning framework from per-step action selection to upfront
    policy authoring. The recursion is structured as a depth-$K$
    call stack rather than a $K$-round outer loop:
    \begin{equation}
    \texttt{KReason}(k)\;=\;\begin{cases}
      \text{LLM}(\text{rules}) & k=1,\\
      \text{LLM}\!\bigl(\text{rules},\;\texttt{KReason}(k-1)\bigr) & k>1.
    \end{cases}
    \end{equation}
    At each level $k$, both seats' policies are jointly authored in a single LLM call conditioned on the level-$(k{-}1)$ profile: the level-$k$ first-seat policy is the best response to th level-$(k{-}1)$ second-seat policy, and the level-$k$ second-seat policy is the best response to the level-$(k{-}1)$ first-seat policy. The recursion bottoms out at $k{=}1$ with a rules-only naive profile, mirroring Algorithm~1 of \citet{zhang2025klevel} with the recursive unit replaced from a predicted action to a written policy profile for the policy generation tasks.
\end{itemize}
For every baseline, generation runs over the same eight seeds, the same 13 scenarios, and the same structured policy style prompt, so all rows in the comparison share an identical evaluation protocol.

\subsection{Computational Resource}\label{app:comp}
The experiments are conducted on a single node with 2 $\times$ NVIDIA A100 (80GB) GPUs. The total computational cost is approximately 300 A100 GPU hours.

\section{Additional Results}

\begin{figure}
    \centering
    \includegraphics[width=1\linewidth]{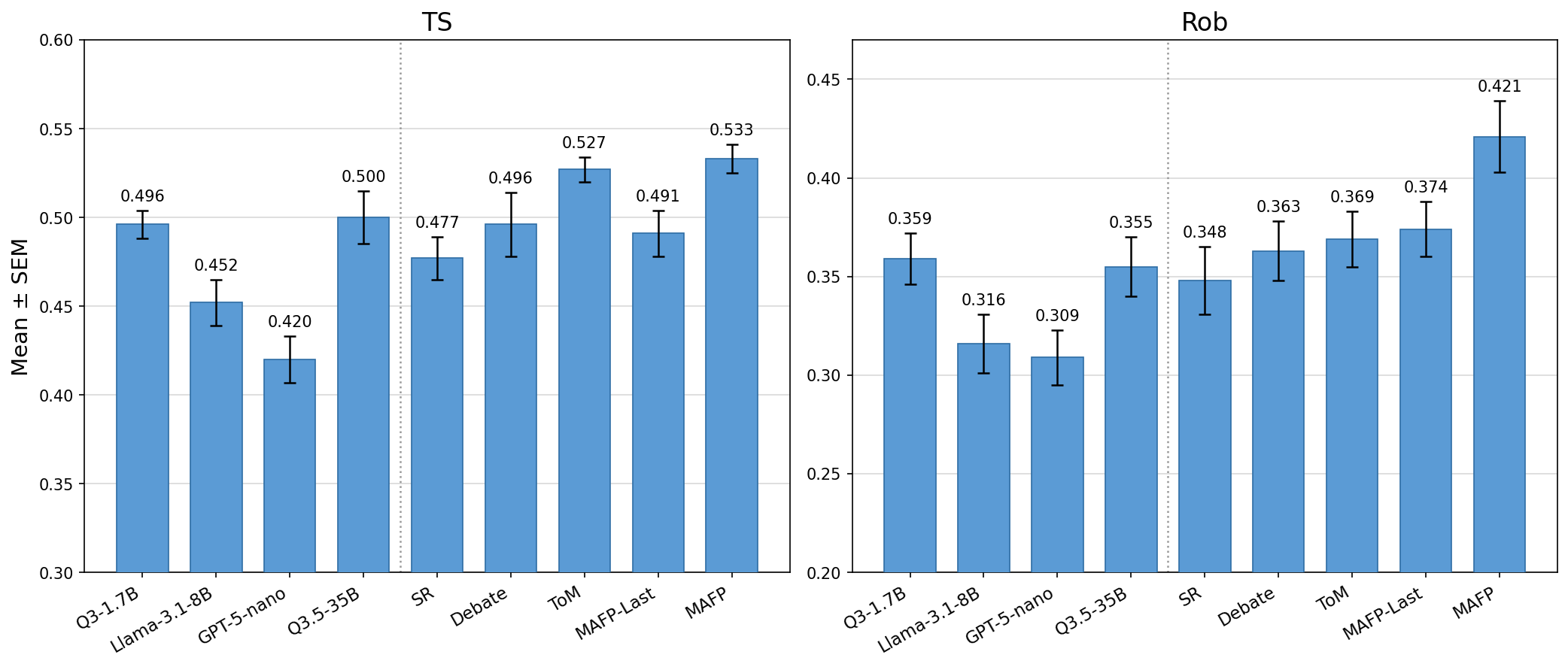}
    \vspace{-2em}
    \caption{Per-method results with error bars visualization.}
    \label{fig:error}
\end{figure}

Here we report the error bars of the main tables in Table~\ref{tab:error} and visualize them in Fig.~\ref{fig:error}. The results show that MAFP performs comparably to ToM on Tournament Strength while clearly surpassing all other baselines, and exhibits a pronounced advantage over every baseline on Robustness, confirming the statistical significance of our experimental conclusions.

\begin{table}[t]
\centering
\caption{Per-method results with error bars (mean $\pm$ SEM).}
\label{tab:error}
\begin{tabular}{l c c}
\toprule
Method & TS & Rob \\
\midrule
\multicolumn{3}{c}{\textit{Single Round}} \\
\midrule
Q3-1.7B           & $0.496 \pm 0.008$ & $0.359 \pm 0.013$ \\
Llama-3.1-8B      & $0.452 \pm 0.013$ & $0.316 \pm 0.015$ \\
GPT-5-nano        & $0.420 \pm 0.013$ & $0.309 \pm 0.014$ \\
Q3.5-35B          & $0.500 \pm 0.015$ & $0.355 \pm 0.015$ \\
\midrule
\multicolumn{3}{c}{\textit{Multiple Rounds with Q3.5-35B}} \\
\midrule
SR                & $0.477 \pm 0.012$ & $0.348 \pm 0.017$ \\
Debate            & $0.496 \pm 0.018$ & $0.363 \pm 0.015$ \\
ToM               & $0.527 \pm 0.007$ & $0.369 \pm 0.014$ \\
MAFP-Last           & $0.491 \pm 0.013$ & $0.374 \pm 0.014$ \\
\textbf{MAFP}       & $\mathbf{0.533 \pm 0.008}$ & $\mathbf{0.421 \pm 0.018}$ \\
\bottomrule
\end{tabular}
\end{table}

\section{Prompt Templates}

We present the prompt templates for the three LLM-based components of our pipeline in Figs.~\ref{fig:prompt-agg},~\ref{fig:prompt-br}, and~\ref{fig:prompt-act}. The aggregation operator $\mathrm{Agg}_M$ aggregates a player's past policies into a single policy that approximates the empirical mixture over the pool. The best-response operator $\mathrm{BR}_M$ authors a policy that maximizes expected win-rate against this averaged opponent. For the action operator $M_{\text{act}}$, we uses GTBench's prompt\_agent template~\cite{duan2024gtbench} and appends the learned policy to the user message.

\begin{figure}[t]
\centering
\begin{promptbox}
\textbf{System prompt}

System Role: You are a strategy summariser. Given $K$ past policies authored
by one player (the opponent) across successive fictitious-play rounds,
produce ONE synthesised policy that best represents the uniform mixture over
that pool. Capture the principles they share, preserve tactical patterns
that recur across iterates, and break idiosyncratic ties neutrally. Do NOT
cherry-pick a single iterate, and do not invent tactics that no iterate
endorses. The output will be handed to a downstream best-response optimiser
as if it were the opponent's actual policy for the next game.

Output Format: only the synthesised policy text. No commentary, no markdown
fences, no mention of iterate numbers. Produce a structured policy with
clearly labelled sections (for example: Opening Principles, Midgame
Priorities, Endgame / Closing Rules, Tactical Checks) where each section
lists concrete, state-grounded rules. Each section header should end with a
colon and appear on its own line. Keep sections focused: the overall policy
should be thorough but not exhaustive --- prefer depth on the key decision
points over enumerating every variation.

\tcbline
\textbf{User prompt}
\begin{flushleft}
Task: $\langle$task$\rangle$\\
Target seat being summarised: $\langle$seat$\rangle$\\
Number of past policies to average: $\langle K\rangle$\\[6pt]
Game rules:\\
$\langle$game rules$\rangle$\\[6pt]
Past policies for the $\langle$seat$\rangle$ seat (treat these as a uniform mixture):\\
{}[Opponent iterate \#1]\\
$\langle$iterate 1$\rangle$\\[6pt]
{}[Opponent iterate \#2]\\
$\langle$iterate 2$\rangle$\\
\quad$\vdots$\\[6pt]
Produce a single synthesised $\langle$seat$\rangle$ policy, organised into
clearly labelled sections, representing the average of the pool above.
\end{flushleft}
\end{promptbox}
\caption{Prompt template for the aggregation operator
$\mathrm{Agg}_M$. $\langle\cdot\rangle$ marks
runtime-filled slots; $\langle$seat$\rangle\in\{$first player, second
player$\}$.}
\label{fig:prompt-agg}
\end{figure}

\begin{figure}[t]
\centering
\begin{promptbox}
\textbf{System prompt}

System Role: You are a game strategy author performing fictitious play in
language space. The opponent's strategy for the next game is the SINGLE
policy shown below (a synthesised average over their past iterates). Author
a playing policy from scratch that maximises expected win-rate against this
average opponent.

Output Format: only the policy text. No commentary, no markdown fences.
Produce a structured policy with clearly labelled sections (for example:
Opening Principles, Midgame Priorities, Endgame / Closing Rules, Tactical
Checks) where each section lists concrete, state-grounded rules. Each
section header should end with a colon and appear on its own line. Keep
sections focused: the overall policy should be thorough but not exhaustive
--- prefer depth on the key decision points over enumerating every
variation.

\tcbline
\textbf{User prompt}
\begin{flushleft}
Task: $\langle$task$\rangle$\\
Your seat: $\langle$seat$\rangle$\\
Opponent seat: $\langle$opponent seat$\rangle$\\
Fictitious-play round: $\langle t{+}1\rangle/\langle T\rangle$\\[6pt]
Game rules:\\
$\langle$game rules$\rangle$\\[6pt]
Opponent's average policy (synthesised from their past iterates):\\
$\langle$averaged opponent policy$\rangle$\\[6pt]
Author a fresh $\langle$seat$\rangle$ policy, organised into clearly
labelled sections, that maximises expected win-rate against the opponent's
average policy above.
\end{flushleft}
\end{promptbox}
\caption{Prompt template for the best-response operator $\mathrm{BR}_M$.}
\label{fig:prompt-br}
\end{figure}

\begin{figure}[t]
\centering
\begin{promptbox}
\textbf{System prompt}

You are a powerful gaming agent who can make proper decisions to beat the user in gaming tasks. You are a helpful assistant that strictly follows the user's instructions.

\tcbline
\textbf{User prompt} (\texttt{prompt\_agent} format; observation $o_t$ = rules $+$ state)
\begin{flushleft}
$\langle$game rules$\rangle$\\
$\langle$current state: moves so far $+$ legal actions$\rangle$\\
You must choose an legal action to set up advantages.\\[6pt]
Your output must be in the following format:\\[6pt]
Action:\\
Your action wrapped with \texttt{<>}, $\langle$action-format spec$\rangle$\\[6pt]
Please return your answer without explanation!\\[6pt]
Learned policy:\\
$\langle$policy text$\rangle$
\end{flushleft}
\end{promptbox}
\caption{Prompt template for the action model $M_{\text{act}}$.}
\label{fig:prompt-act}
\end{figure}

\end{document}